\documentclass[preprint,12pt]{elsarticle}

\usepackage{amssymb}

\usepackage{times}
\usepackage{epsfig}
\usepackage{graphicx}
\usepackage{amsmath}
\usepackage{amssymb}
\usepackage{subfigure}
\usepackage{bm}
\usepackage{tabularx} 
\newcolumntype{Y}{>{\centering\arraybackslash}X}

\usepackage{array,booktabs,arydshln,xcolor} 
\usepackage{amsthm}
\usepackage{epstopdf}
\usepackage{algorithm}
\usepackage{amsthm}
\usepackage{multirow}
\usepackage[algo2e]{algorithm2e}
\usepackage[breaklinks=true,letterpaper=true,colorlinks,bookmarks=true]{hyperref}
\usepackage{bbding}
\usepackage{pifont}
\usepackage{wasysym}
\usepackage{amssymb}
\usepackage{tabularx}
\usepackage{rotating}

\journal{arXiv}

\begin{document}

\begin{frontmatter}



\title{Recent Advances in the Applications of Convolutional Neural Networks to  Medical Image Contour Detection}


\author[label1]{Zizhao Zhang\footnote{E-mail: zizhaozhang@ufl.edu}}
\author[label1]{Fuyong Xing}
\author[label1]{Hai Su}
\author[label1]{Xiaoshuang Shi}
\author[label1]{Lin Yang }

\address[label1]{University of Florida}

\begin{abstract}
The fast growing deep learning technologies have become the main solution of many machine learning problems for medical image analysis.
Deep convolution neural networks (CNNs), as one of the most important branch of the deep learning family, have been widely investigated for various computer-aided diagnosis tasks including long-term problems and continuously emerging new problems. Image contour detection is a fundamental but challenging task that has been studied for more than four decades. Recently, we have witnessed the significantly improved performance of contour detection thanks to the development of CNNs. Beyond purusing performance in existing natural image benchmarks, contour detection plays a particularly important role in medical image analysis. Segmenting various objects from radiology images or pathology images requires accurate detection of contours. However, some problems, such as discontinuity and shape constraints, are insufficiently studied in CNNs. It is necessary to clarify the challenges to encourage further exploration.
The performance of CNN based contour detection relies on the state-of-the-art CNN architectures. Careful investigation of their design principles and motivations is critical and beneficial to contour detection. In this paper, we first review recent development of medical image contour detection and point out the current confronting challenges and problems. We discuss the development of general CNNs and their applications in image contours (or edges) detection. We compare those methods in detail, clarify their strengthens and weaknesses. Then we review their recent applications in medical image analysis and point out limitations, with the goal to light some potential directions in medical image analysis. We expect the paper to cover comprehensive technical ingredients of advanced CNNs to enrich the study in the medical image domain.
\end{abstract}


\end{frontmatter}

\section{Introduction}
\label{sec:introduction}
Medical image analysis is the foundation of a computer-aided diagnosis (CAD) system. The analysis of medical images with different modalities usually requires accurate segmentation to isolate abnormal objects (cells or organs) to support efficient quantization. Contour detection is a fundamental prerequisite for medical image segmentation, with the aim to detect the edges from images and further collect the knowledge of the contours of objects. Accurate and fast contour detection is a long-term study in this domain, which suffers from many difficulties. In recent years, we have witnessed inspirational renovation in medical image analysis, particularly image segmentation, due to the development of advanced machine learning technologies.

Edge is the basic components of images. Detecting object edges and contours is critical in many practical computer vision and medical image computing tasks. The research in contour detection is a huge family comprised by a large number of directions using various computer vision, image processing, and machine learning techniques \cite{papari2011edge}. Early contour detection methods are dominated by unsupervised approaches, with the aim to estimate the local gradient changes. In the past five years, supervised methods gradually dominate this area as a result of both accuracy and efficiency advantages. The main idea is to train a machine learning classifier to predict central pixel labels (edge or non-edge) of local patches. Both directions require heavy hand-crafted features to accurately represent the local gradient information. 
Contour detection includes the detection of edges but can simultaneously outline the continuous edges belong to object contours. Therefore, contour detection is substantially more challenging than edge detection since it models both low-level gradients and high-level object information.
Under conventional directions, the integration of low-level and high-level cues is difficult, which often results in complex and computationally demanding frameworks, including pre-processing, feature engineering, classifier training, and post-processing. A fast and highly-integrated method that can accept raw images and output contour maps is quite hypothetical, and the advantages of deep learning give hopes to the demand. 

There is world-wide recognition that deep learning has advanced the artificial intelligence (AI) to the next generation \cite{lecun2015deep,Goodfellow-et-al-2016,schmidhuber2015deep}. 
The family of deep learning is comprised of a number of unsupervised and supervised learning models \cite{hinton2011deep,lee2009convolutional,lee2009unsupervised,goodfellow2014generative}, such as Restricted Boltzmann Machine (RBM) \cite{salakhutdinov2007restricted,nair2010rectified,salakhutdinov2009deep} and Recurrent Neural Networks (RNNs). CNN is a supervised model widely used for image understanding.
Compared with conventional machine learning models, such as support vector machines (SVM) \cite{cortes1995support}, random forests \cite{liaw2002classification}. CNN has a deep layer-wise structure, making it proficient at learning hierarchical and nonlinear representations to represent and discover complex and intricate high-dimensional data structures to support discriminative classification. Therefore,  CNN is also a representation learning method \cite{bengio2013representation}.

In edge detection, there are multiple remarkable studies using standard CNNs \cite{shen2015deepcontour,ganin2014n,bertasius2014deepedge} have achieved marginal improvement over conventional methods. However, the standard CNN suffers from the computationally bottleneck of their patch-to-pixel dense prediction paradigm. 
The development of contour detection continuously takes benefits from the development of semantic segmentation \cite{noh2015learning,chen2014semantic,hariharan2015hypercolumns,dai2015boxsup,lin2015efficient,xie2015holistically,pathak2014fully,yang2016object,maninis2016convolutional}. One of the most critical techniques for dense prediction tasks is end-to-end CNN, proposed by \cite{long2015fully,long2016fully}, which performs pixel-wise prediction in a single feedforward. At present, end-to-end training becomes the standard for most kinds of structured outputs, such as bounding boxes \cite{ren2015faster,girshick2014rich}, shape \cite{jaderberg2015spatial}, orientations \cite{maninis2016convolutional}, which offers much flexibility to CNN beyond the form of image labels. The end-to-end training manner for dense pixel prediction \cite{xie2015holistically,yang2016object,maninis2016convolutional} dramatically improves the performance of contour detection, surpassing a significant margin over preceding standard CNN based methods and even surpassing the empirical accuracy of human annotators.
These advantages dramatically affect the medical image domain.

Different from natural images, medical images do not have rich semantic information, effective usage of low-level edge information is the key to support accurate segmentation. Moreover, the failure of detecting edges will cause huge issues in diagnostic precision, for example, the failure of segmenting touching cells will cause abnormal cell size statistics.
Therefore, contour detection is usually treated as an intermediate step for image segmentation. 
Some segmentation studies have implicit contour detection because the main challenging of object segmentation is the accurate location of edges (this paper will take such kinds of segmentation methods into the consideration), such as segmenting neuronal
membranes in electron microscopy images \cite{ciresan2012deep} and vessels in retinal images \cite{ganin2014n}. 
We have witnessed numerous state-of-the-art CNN based methods being successfully applied to medical image contour detection and segmentation \cite{chen2016dcan,maninis2016deep,ronneberger2015u}. However, direct technical transferring sometimes conceals several critical problems in the medical images but may not being concerned in the natural image domain, such as the detection of weak edges, the processing of discontinuity of contours, and the detection of edges from high signal-to-noise ratio (SNR) images. The solutions to these problems are significant in medical image analysis.

We start by discussing the difficulties of edge/contour detection in medical image computing and review some conventional approaches in Section \ref{sec:mededgeoverview}
To better understand the development of CNNs in image contour detection, we briefly introduce the principle of CNNs in Section \ref{sec:cnnarch} and discuss most state-of-the-art CNN based methods for edge detection in Section \ref{sec:edgedet}, with the goal to clarity the key problems they are addressing and their advantages for medical image usage. After understanding the principle of CNNs for contour detection, in Section \ref{sec:meidcaledgedet}, we review recent methods for medical image contour detection and segmentation to help understand underlying technical basics of current methods in medical image analysis and build the connection to the state-of-the-art method in the computer vision community, and also show the limitations of current methods.
Section \ref{sec:discussion} discusses some key problems and potential directions. Section \ref{sec:conclusion} concludes the paper.
We expect this paper can cover the necessary technical advances in CNNs that is useful for contour detection and, more importantly, can attract attentions of the underlying problems and lead to further exploration of the CNN technologies in medical image analysis. 

%

\begin{figure}[t]
	\begin{center}
		\includegraphics[width=0.9\textwidth]{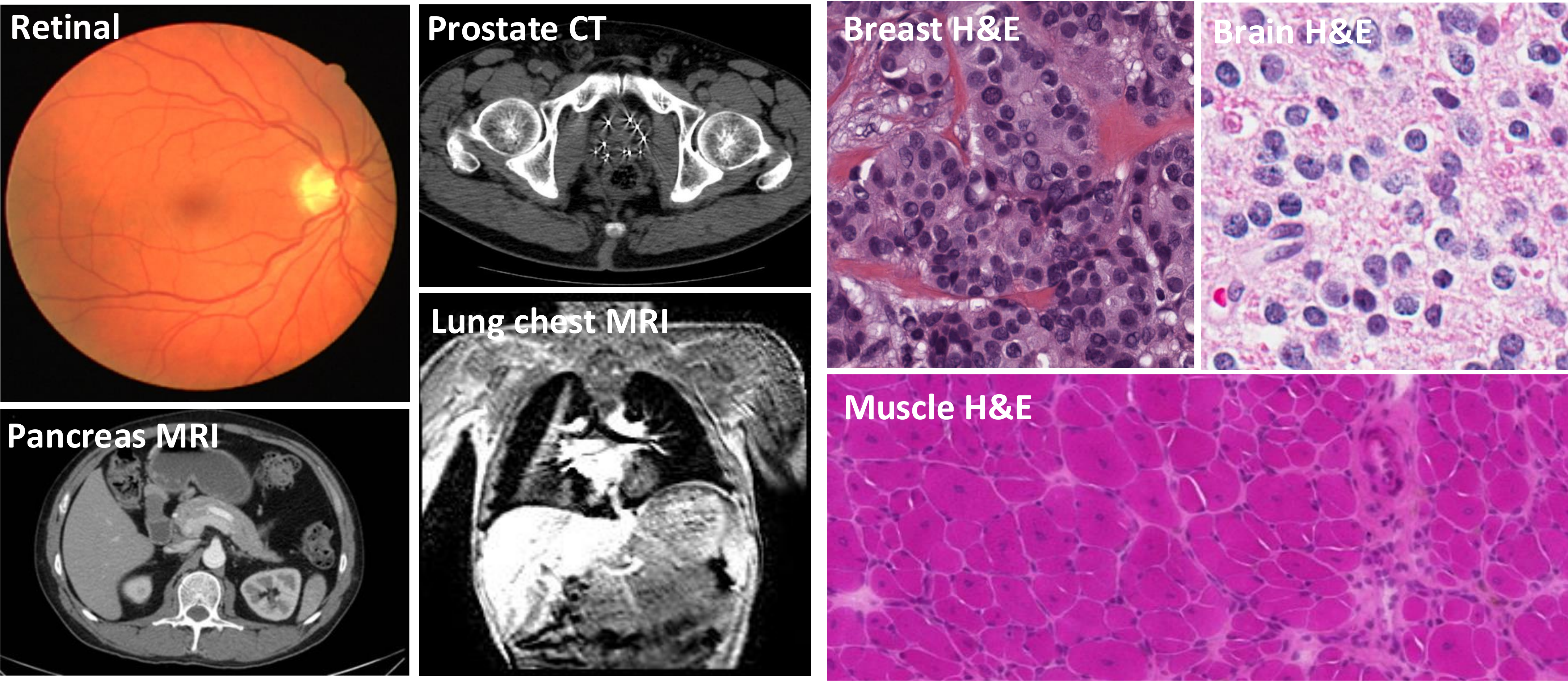}
	\end{center}
	\caption{The left side shows four kinds of CT, MRI, or tomography organ images. The right side shows three kinds of microscopic images. Segmenting the objects (e.g. pancreas or nucleus) needs clear detection of object contours. In breast or muscle images, detecting contours is obviously very challenging due to the severe touching objects and artifacts.} \label{fig:medimg}
\end{figure}

\section{Overview of Contour Detection for Medical Images}
\label{sec:mededgeoverview}
In this section, we provide an overview of the challenges and significance of image contour detection in medical images. Then we review several kinds of conventional directions for medical image contour detection, with the goal to encourage inspirations in CNN designing.

\subsection{Challenges and significance}
{Compared with natural images containing all kinds of semantic objects,
	medical images are more modality-specific such that in one modality, there is less semantic and texture information inside or between objects. In radiological data like pancreas MRI or CT or ultrasound images, the targeting objects are organs or bones. In pathological images like hematoxylin and eosin (H$\&$E) stained lung cell specimens \cite{el2013computer,xie2015beyond, madabhushi2016image}, the targeting objects are cells and diseased regions. Organs and cells usually have consistent appearance. Therefore, object shapes and structures play a key role for medical image segmentation or detection. Figure \ref{fig:medimg} shows some types of medical images that need careful process of object contours to achieve accurate segmentation.}

The image quality defection due to the acquisition and imaging processes is common and significant \cite{suzuki2003neural,6191314}. Noises bring obstacles to edge detection because it reduces the contrast of real edges and also introduce spurious edges due to noisy contrast. Although applying denoising algorithms before some local edge detection can reduce the effects to some extent, this approach has been shown not very promising \cite{ofir2015fast}. A global method with some prior knowledge of targeting objects is supposed to overcome the effects of local noises. The fine detection of edges and global object contours are equally important and require discrimination when prediction. For example, in retinal images, the detection and segmentation of blood vessels require very fine detection of subtle edges. While in pathological images, ignoring gradients caused by staining noises and boundaries of small cells is critical.

{Detecting weak or broken edges due to occlusion or staining artifacts between touching or overlapping objects is a long-term studied problem in medical image analysis. Sometimes detecting these kinds of edges which are even visually in-discriminative seems impossible, but learning to link the broken edges is a remedial measure \cite{hajjar1997new}. We believe this problem is an active research topic of the contour detection. In the earlier stage, deformable models \cite{kass1988snakes,chan2001active,xu1998snakes,caselles1997geodesic} are popular techniques to guarantee the continuity and smoothness of object contours, because active contour models use a parametric or non-parametric representation to initialize object boundaries and transform these closed contours to align with objects. However, currently this area is not active because active contour is built on particular assumptions. Recently, we have seen work \cite{rupprecht2016deep} that train CNNs to predict the movement of active contours.}

Besides direct contour detection, \cite{wang2007global,stahl2008globally,levinshtein2010optimal} have studied how to detect global, closed, or convex object contours from a set of broken contour pieces containing interesting edges and noisy edges. However, these methods are difficult to generalize to real datasets due to the relied assumptions, such as bilateral symmetry of objects. Additionally, \cite{ren2005scale} have studied contour completion using conditional random field (CRF) models, and \cite{liu2014touching} extends the technique to handle medical images. \cite{yangshape} and \cite{su2015robust} use RNNs and autoencoders \cite{hinton1994autoencoders} to achieve contour completion.

High-level reasoning is an important factor for contour completion.
Human can easily recognize the broken edges between objects and identity touching or overlapping objects, although the actual gradient in the broken edges is hardly seen. We believe there are two reasons at least \cite{shapley1973edge,sanguinetti2013ground,hsieh2010recognition,kourtzi2001representation}:
\begin{enumerate}
	\item Human vision has a strong reasoning ability by observing surrounding object contours connecting to the broken points, and thus it can easily recognize and predict the existence of broken edges.
	\item Human vision has high-level prior knowledge about the observing objects' appearance \cite{kourtzi2001representation}, so it can estimate rough object appearance and reject unfamiliar appearance (touched and connected multiple objects).
\end{enumerate}

Both need to take advantage of the context information. However, as we previously discussed, most of previous CNN based contour detection and segmentation methods do not focus on this kind of context information. In semantic segmentation tasks, we have noticed a lot of work to model the context information using methods like CRF and markov random field (MRF) \cite{lin2015efficient,liu2015crf,alvarez2012semantic,li2016combining,chen2014semantic,zheng2015conditional,chen2014semantic,liu2015semantic}. The inference of edges is supposed to be more difficult because it contains less semantic knowledge and require more complex understanding of shapes and structures of objects. For example, \cite{fuocclusion} have studied occlusion boundary detection by exploring deep context from CNNs, including local structural boundary patterns, observations from surrounding regions, and temporal context.

In addition, the effective learning from limited annotated data is an significant topic because of the difficulty in collecting large-scale medical image datasets. Usually CNNs require large-scale datasets to train. Semi-supervised and unsupervised learning methods \cite{li2015unsupervised} and transfer learning \cite{shin2016deep,tajbakhsh2016convolutional} are recently been discussed in the study of CNNs. In addition, a better CNN architecture design can significantly increase the efficiency of parameter utilization. We will discuss the details in the following.

\subsection{Previous medical image contour detection}
Conventional contour detection and segmentation methods before the prevalence of deep learning have numerous directions in medical image computing, and a comprehensive review can be found in \cite{xing2016robust}.

Intensity thresholding \cite{Gonzalez08} is one of the early-stage approaches for medical image segmentation. For some specific image modalities, such as fluorescence microscopy images, where the target objects (e.g., nuclei or cells) are usually brightly stained in a dark background, it is effective to apply intensity thresholding to object segmentation \cite{Chen06thresh}; however, it is difficult for thresholding to handle other image modalities (e.g., H\&E stained images), especially for touching or partially overlapped objects.

Watershed transform \cite{roerdink2000watershed} is a popular segmentation method in medical images \cite{6191314}, which pursues the `catchment basins' (gradient discontinuity points). It can be used to find the continuous contours of objects, and therefore it is popular at segmenting multiple objects like cells in pathological images \cite{Meijering12}. However, watershed usually suffers from over-segmentation, and thus marker-controlled watershed \cite{Gonzalez08} has been proposed for effective contour detection and segmentation. An alternative method to handle over-segmentation is to merge falsely segmented regions with certain criteria \cite{Lin05ws}.

One widely-studied direction for medical image contour detection and segmentation is deformable models \cite{Delgado15dm}. Deformable model based methods focus on deforming an initial active contour to align the object boundary by solving an energy function. Representative deformable models include geodesic models or level-set models (such as Chan-Vese model \cite{chan2001active}) \cite{caselles1997geodesic}, parametric models (Snake \cite{kass1988snakes} and GVF \cite{xu1998snakes}). Many related work has been proposed for object contour delineation \cite{Xing14detcnt,zhang2004tracking,li2008cell,cremers2007review} and some of them are combined with shape prior modeling for touching object segmentation in medical images \cite{Ali12lsshp,Xing16detseg}. One potential limitation of deformable models is the requirement of proper contour initialization, which might be achieved using effective object detection methods \cite{xing2016robust}.

Graph-based methods are another popular category of methods for medical image contour detection and segmentation. In graph partition, the max-flow/min-cut algorithm \cite{Boykov01,Boykov04} is usually used to minimize an energy function, and it has been successfully applied to contour detection in medical images \cite{Kofahi10,Chang13}. Normalized cut \cite{shi2000normalized} is proposed to avoid the bias of favoring small sets in the global minimum cut, and a generalized normalized cut \cite{Bernardis10} has been proposed for object segmentation in microscopy images. Some other graph partitioning methods such as random walk \cite{Grady06rw} and isoperimetric partitioning \cite{Grady06} have been also reported for object segmentation in medical image data.

Conventional machine learning methods have been applied to medical image contour detection and segmentation. For pixel-wise classification-base segmentation, it is usually necessary to conduct further processing to split touching objects \cite{Kong11}; for superpixel-wise classification, it would improve the computational efficiency, but the pre-generated superpixels need to well adhere real object boundaries \cite{Janssens13sl}. In addition, the conventional machine learning approaches require manual feature representation design, which is not a trivial task in some medical applications.

\begin{figure}[t]
	\begin{center}
		\includegraphics[width=0.9\textwidth]{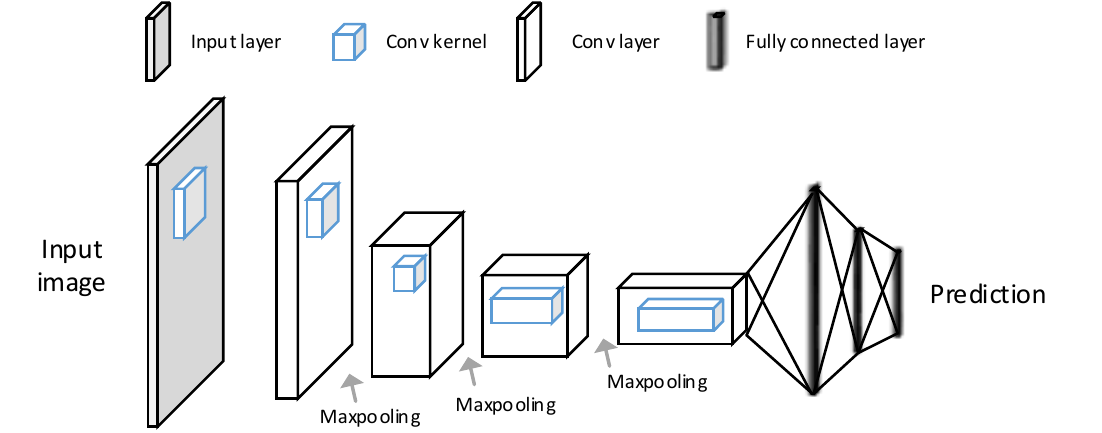}
	\end{center}
	\caption{An illustration of the architecture of a CNN, including five convolutional layers, three max pooling layers, and two three fully connected layers. The first convolutional layer takes an input image and the last fully connected layer predicts the label. } \label{fig:outline}
\end{figure}

\section{Convolutional Neural Networks}
\label{sec:cnnarch}
In this section, we briefly introduce the basic concept of CNNs.
As the predecessor of CNN, ANN is originally inspired by the goal of modeling biological neural systems. Its organization simulates the mechanism of information transmission in the brain neuron. The computation unit contains a linear transformation $\bm z_i = \sum_i w_i \bm x_i + b $ plus an activation function $\bm y_i  = \frac{1}{1+e^{-\bm z_i}}$ (e.g. Sigmoid) on the input $\bm x$ and generate an output $\bm y$. $\bm w = [w_1,...,w_n]^T$ is the weights function connecting previous neurons to next neutron. Computation units are connected one after another, which compose a layer-wise organized architecture. 
When connecting to the output layer, a Softmax or Sigmoid function is often used to map the output to $[0,1]$, representing label probabilities. 
The above formulation is a very basic ANN design, which primarily models the biological neuron system. After that, the design of ANN towards to the machine learning and engineering guidance.

\subsection{Architecture}
CNN has a very similar architecture with ANN, which is composed by a series of computational layers \cite{lecun1989backpropagation}. Each layer contains a set of neurons and may have learnable parameters to decide how the neutrons between layers are connected. Therefore, the overall architecture is structured by cascaded layers.  Figure \ref{fig:outline} shows a CNN architecture with eight layers.

The main building brick is the convolutional (Conv) layer, whose Conv kernels (or filters) perform convolutional operation across the whole image spatial locations to generate output image  representations (i.e. feature maps). The spatial extent of kernels refer to as the receptive field. This local connectivity through the receptive field is originally inspired by the brain neuron science \cite{duhamel1997spatial}. Pooling layer is inserted between Conv layers for the purpose of downsampling the feature map dimension. The most common used pooling layer is max pooling, which keeps the highest response value in an image extent and discard the rest, and perform this operation crossing the whole image.
Activation layers map data points non-linearly. The appropriate settings of activations is critical to the behaviors of CNN training.
The most common used activation at present for CNN is ReLU \cite{nair2010rectified}, which simply performs $y = \max(0,x)$. It is applied after a Conv layer or a fully connected layer. ReLU is an important technique for modern CNNs. Previous activation functions such as Sigmoid and Tanh suffer from strong gradient vanishing (or gradient saturation) problems \cite{glorot2011deep}, which can be accumulated and getting severer as the layer increases.

Apart from the basic layers, there are a variety of layers proposed by modern CNNs. For example,  Dropout \cite{srivastava2014dropout}  and batch normalization \cite{ioffe2015batch} are now standard configurations in the CNN design patterns. Modified/generalized convolutional layers \cite{yu2015multi,long2015fully}, ReLU layers \cite{maas2013rectifier}, and pooling layers \cite{zeiler2014visualizing}  are also specifically designed for various applications.  We will discuss some of them in the following.

\subsection{Network training}
Recently, the improvement of the optimization algorithm, hardware capacity, and functional deep learning libraries, make the training more easier than before. A beginner with moderate experiences can deploy the training of very deep networks in a few lines of code on a GPU. The main component of CNN training is stochastic gradient descent (SGD) \cite{bottou2010large} and backpropagation \cite{lecun2012efficient}.

SGD minimizes the empirical risk of the loss function $J(\theta)$ by updating the parameters $\theta$:
\begin{equation}
\label{eq:learning}
\bm \theta_{t+1} = \bm \theta_{t} - \gamma \sum_{1}^{m}  \nabla_{\theta} E[J(\theta)]
\end{equation}
where $n$ is the number of observed training data and $\gamma$ is the learning rate. SGD  randomly sample a mini-batch of $m$ samples from total training samples and update the parameters using the averaged gradients generated by mini-batch samples.
Standard SGD could easily trap at local optima and lead to slow convergence \cite{sutskever2013importance}.
Then momentum method is introduced to resolve this problem by controlling the gradient velocity during optimization \cite{sutskever2013importance}. The SGD with momentum is defined as
\begin{equation}
\bm v_{t+1} = \lambda  \bm v_{t} - \gamma \sum_{1}^{m}  \nabla_{\theta} E[J(\theta)] \
\end{equation}
\begin{equation}
\bm \theta_{t+1} = \bm \theta_{t} - \bm v_{t+1}
\end{equation}
where $\lambda\in[0,1]$ is the momentum coefficient. Based on the basic SGD with momentum, there are also new algorithms to improve the training efficiency and lead to better convergence, such as Nesterov momentum \cite{sutskever2013importance}, Adagrad \cite{duchi2011adaptive},  RMSprop \cite{hinton2012neural}, Adam \cite{kingma2014adam}. Selecting a appropriate learning rate is tricky. The last three can adaptively adjust the learning rate per parameter at each update, which have been shown to lead to faster convergence.
Current state-of-the-art methods still use different optimization algorithms based on specific applications. The optimal choice depends on specific problems.

Backpropagation propagates the errors computed in the loss layers back to all proceeding layers. Every computational layer will generate the gradient w.r.t. its own parameters accordingly. The overall procedure follows the basic chain rule. Let's define the object function $J$ as
\begin{equation}
J(\bm x,\bm g;\theta) = \frac{1}{m} \sum_{i=1}^{m} loss(\bm g_i, f(\bm x_i;\theta)),
\end{equation}
where $loss$ computes the difference between the prediction $f(x_i;\theta)$ and the groundtruth $\bm g_i$. There are various types of loss functions, such as Cross-entropy, Softmax, Euclidean distance, Max-margin, etc, for botch classification and regression purposes. $f$ denotes overall function of CNN, so $f$ takes an input image $x$ and outputs $y_L$ as its predicted label. Suppose $y_L$ is computed by a fully connected layer (the $L$- layer of the network), defined as
\begin{equation}
y_L = \sigma (W_L y_{L-1} + b_L)
\end{equation}
where $y_{L-1}$ is the output of the $(L-1)$-th layer. $\sigma$ is the activation function. The gradient of $J$ w.r.t $W_L$ and $y_{L-1}$ is defined as:
\begin{equation}
\label{eq:gradient}
\frac{\partial J}{\partial W_L} = \frac{\partial J}{\partial y_L} \cdot  \frac{\partial y_L}{\partial \sigma} \cdot \frac{\partial \sigma}{\partial W_L} ,
\end{equation}
\begin{equation}
\label{eq:gradient2}
\frac{\partial J}{\partial y_{L-1}} = \frac{\partial J}{\partial y_L} \cdot  \frac{\partial y_L}{\partial \sigma} \cdot \frac{\partial \sigma}{\partial y_{L-1}}
\end{equation}
Eq. (\ref{eq:gradient}) computes the gradients respecting to the weights of layer $L$ and Eq. (\ref{eq:gradient2}) computes the gradients respecting to the input $y_{L-1}$ of layer $L$.

Successfully training CNN networks is not as simple as its mathematical definition. The overall optimization is highly non-convex and the process is difficult to visualize. Overfitting is one of the long-term challenge actively studied in the community.
There are a wide range of well-investigated approaches that substantially alleviate CNN training difficulties, for instance, data augmentation, weight initialization \cite{hinton2006reducing,sutskever2013importance,glorot2010understanding,he2015delving}, regularizations \cite{srivastava2014dropout,wan2013regularization}, activations \cite{nair2010rectified,glorot2011deep,goodfellow2013maxout,maas2013rectifier,clevert2015fast}, normalization \cite{ioffe2015batch}, and skip-connection \cite{he2015deep}. We refer readers to \cite{gu2015recent} for more details.

\section{State-of-the art CNN Architectures}
\label{sec:sotaCNN}
\begin{figure}[t]
	\begin{center}
		\includegraphics[width=0.8\textwidth]{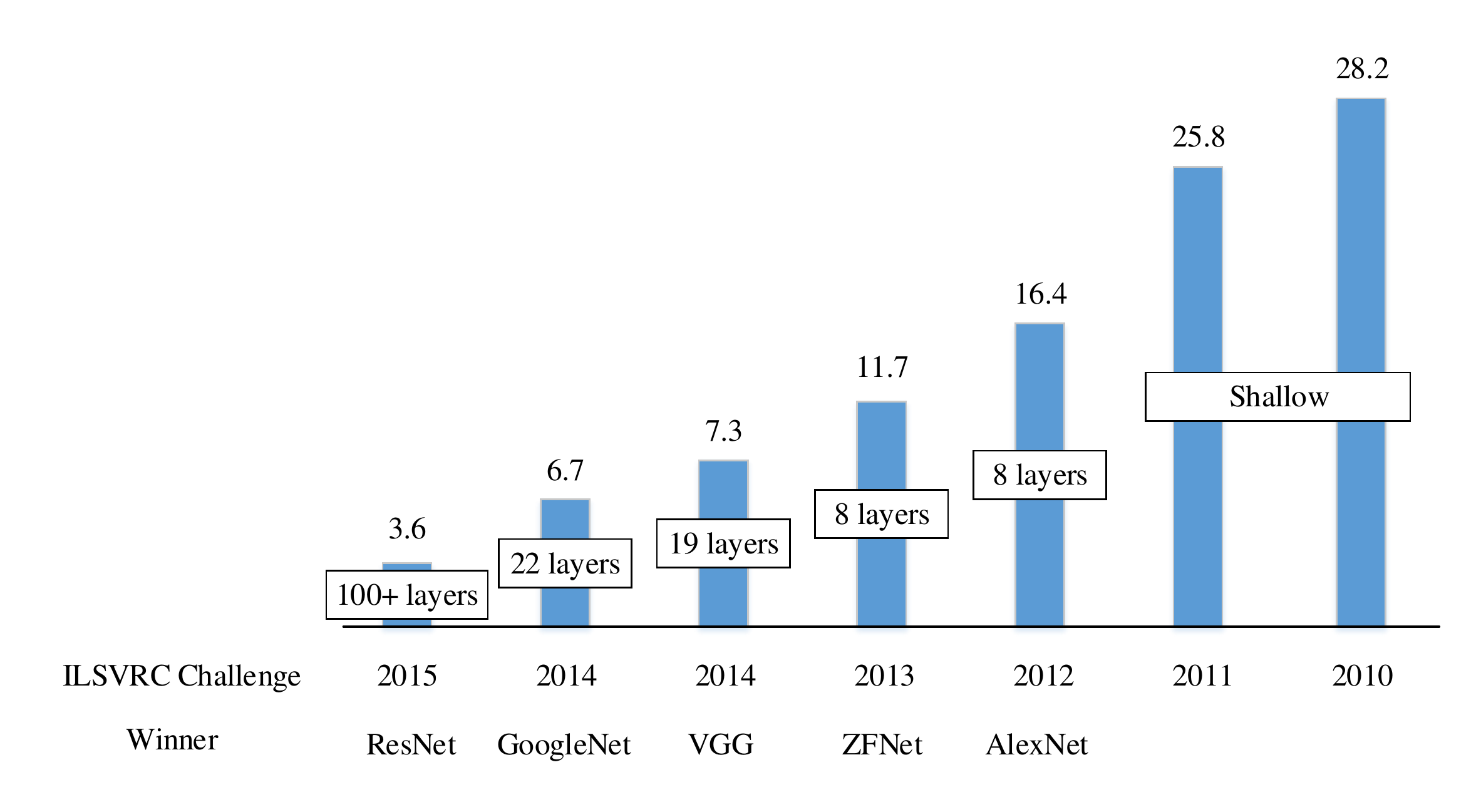}
		
	\end{center}
	\caption{The ImageNet ILSVRC challenge results (top-5 error ($\%$))) from 2010 to 2015. AlexNet achieves a very large margin improvement with deeper network than before. The number of layers is continuously increasing as the accuracy increases. ResNet conquers the barrier of training network over 100 layers, much deeper than previous winners. } \label{fig:challenge}
\end{figure}

In this section, we introduce several well-known CNN architectures, which are recognized as the milestone in the CNN development and the basement of various computer vision tasks, specially image edge detection. We also discuss the related variations to address specific problem in CNNs, such as ensembling, generalization, etc.
The performance of these CNNs is publicly validated on the annual ImageNet Large Scale Image Recognition Challenge (ILSIRC). Figure \ref{fig:challenge} shows the winner networks of ImageNet challenges in the past five years. From 2010 to 2015, the classification error rate has decreased by more than nine times. 


\subsection{CNN architecture benchmarks}

\vspace{.2cm}
\noindent
\textbf{LeNet} stands for the first successful application of CNN. It is proposed by \cite{lecun1989backpropagation} and used for hand-written digit recognition. ConvNet to allow the weight sharing between neurons, which dramatically decreases the heavy parameters needed in ANNs. This network is much shallower compared with recent architectures, including $2$ Conv layers with an intermediate subsampling layer and $2$ fully connected layers.  The main contribution of this work is the usage of local connection to replace the fully connection between network neurons of conventional ANNs.

\vspace{.2cm}
\noindent
\textbf{AlexNet} is recognized as the first deep CNN which is successfully applied onto large-scale image recognition, which is proposed by \cite{krizhevsky2012imagenet}. It won the 2012 ILSIRC. AlexNet has a quite similar architecture with LeNet but has more Conv layers. AlexNet has some better solutions to prevent the overfitting and gradient vanishing problem. First, AlexNet uses the ReLU \cite{nair2010rectified} activation to replace Sigmoid. Second, it applies local response normalization (LRN) scheme before each ReLU to further prevent gradient vanishing effect. LRN is an another way to normalize the data for model generalization. Basically, it normalizes the response value at each receptive fields (divided by the sum of the same spatial location), which force the response value to be in a relative small range.
Third, it uses overlapped pooling kernels. General max pooling applies non-overlapping kernels subsample the feature map. Overlapped pooling is just changing the kernel size larger than stride. The paper argues that overlapping kernels is helpful to prevent overfitting. Fourth, AlexNet uses Dropout to prevent overfitting.
Overall, AlexNet has $5$ Conv layers and $3$ fully connected layers with totally 60M parameters (dominated by the three fully connection layers). This number is very large compared with recent CNNs. 

\vspace{.2cm}
\noindent
\textbf{ZFNet} is proposed by Zeiler and Fegus \cite{zeiler2014visualizing}, which won the 2013 ILSIRC. This network has very similar architecture with AlexNet but more detailed hyperparameter tunning and smaller kernels of bottom Conv layers. ZFNet has 75M parameters. The main contribution of this paper is unpooling and deconvolution, which enable the visualization of the hidden layers. 
{Unpooling and deconvolution are quite novel and `uncommon' combination in CNNs. The usage of these two layers can project deep feature maps to the image space, so as to deeply visualize image features highlighted. Nowadays it has become a very popular research topic \cite{sharif2014cnn,chatfield2014return,oquab2014learning,mahendran2015understanding}.

\vspace{.2cm}
\noindent
\textbf{VGGNet} \cite{simonyan2014very} is very popular network not only in image classification but also in many other applications \cite{xie2015holistically,yang2016object,maninis2016convolutional}. It obtains the second best results in ILSIRC 2014, right behind GoogleNet. Its architecture is quite neat and unique compared with GoogleNet. It has five sets of Conv units. All feature maps in each unit has the same dimension and number. Units are connected by max-pooling layers. VGGNet has a 16-layer and a 19-layer version. 
VGG has $140$M parameters. Thanks to its clean and regular architecture and available pre-trained models, VGG gives researches flexibility to manipulate to internal layer representations. Thus it is the mostly widely-used architecture for high-level computer vision tasks, such as semantic segmentation and edge detection.

\vspace{.2cm}
\noindent
\textbf{GoogleNet} is proposed by \cite{szegedy2015going}. It won the ILSIRC 2014 and largely outperforms AlexNet. Moreover, GoogleNet only has 12X fewer parameters than AlexNet yet much deeper (22 layers). The main contribution of this paper is the introduction of the \textit{Inception module}, with the aim to better utilize the representations in network layers. 
The basic \textit{Inception module} takes an input from an layer and pass to multiple different and independent layers (such as $3\times3$ Conv layer, $5\times5$ Conv layer, max pooling layers) in parallel, then the layers' output are merged together. Different kernel sizes capture multi-scale information. The inception module has been extended as four progressive versions.

Inception V2 \cite{ioffe2015batch} introduces batch normalization.
Inception V3 \cite{szegedy2015rethinking} discusses and summarizes the design principles of the inception module in detail. For instance, factoring $5\times5$ kernels to two small $3\times3$ kernels increases computational efficiency and training speed.
Inception V4 \cite{szegedy2016inception} is the latest version (including multiple variants). This study braces the idea of residual networks (ResNet) \cite{he2015deep} into their design. It also introduces the residual scaling to prevent the instabilities when number of filters exceed 1000.

GoogLeNet uses a global average pooling (averaging the value in the window) to transforms the feature maps of the last Conv layer to a vector and only one fully connected layer is used to perform prediction. Average pooling layer has been recognized as a good alternative to fully connected layer after the last Conv layer since it saves the majority of parameters coming from the fully connected layer and has intrinsic regularization effects for modal generalization. This configuration is first applied by Network in Network (NIN) \cite{lin2013network}, another interesting network architecture which builds multilayer perceptron between Conv layers to enhance the local region information abstraction. Most recent CNNs use this configuration as the classification module.

\vspace{.2cm}
\noindent
\textbf{ResNet}  is the most successful CNN in recent two years, developed by \cite{he2015deep,he2016identity}. It is the first CNN that overcomes the barrier of training networks with more than 1,000 layers.  ResNet won the 1st place in ImageNet classification, detection, localization, COCO detection and COCO segmentation. 
The idea of ResNet has been largely extended and widely used in image classification \cite{veit2016residual,huang2016deep,huang2016densely,targ2016resnet,rastegari2016xnor,szegedy2016inception,shah2016deep}, contour detection \cite{maninis2016convolutional}, object detection \cite{liu2015ssd}.

We have a common understanding that deeper network can give rise to better abstraction, but when depth increases to some level, extra layers will hurt the performance. The initial motivation of ResNet is raised by a common question: why it is difficult to train very deep networks?

For examples, suppose we can train a $30$ layer network, when the depth increases to $30+10$. The error rate will rise up \cite{he2015convolutional,srivastava2015highway}. However, intuitively, if we setting the extra $10$ layers as identity mapping, i.e., passing the same output of $30$-th layer the next $10$ layers. The error rate should be the same. However, this simple identity mapping operation seems difficult for CNN to learn directly. So the author argues that the difficult of training very deep network is due to optimization issue but not architecture itself. To overcome this difficult, ResNet suggests to instead allow CNNs to learn residual mapping, which is expected to be easier than learning identity mapping. 

Let's define the ideal underline mapping as $\mathcal{H}(x)$. ResNet lets the computational units (convolutions) model the mapping of $\mathcal{F}(x) = \mathcal{H}(x) - x$. Therefore, the targeting mapping becomes $\mathcal{F}(x) + x$. This formulation is converted to a novel architecture computational unit with skip-connection, as illustrated in Figure \ref{fig:resnet}. This unit is called ``residual unit''. The overall network is constructed by stacking such computational unit.
The concept of skip-connection in neural networks stems from \cite{raiko2012deep} and Highway Network \cite{srivastava2015highway}, which acts like a gate function to selectively allow the information pass to the following layers.

\begin{figure}[t]
	\begin{center}
		\includegraphics[width=0.5\textwidth]{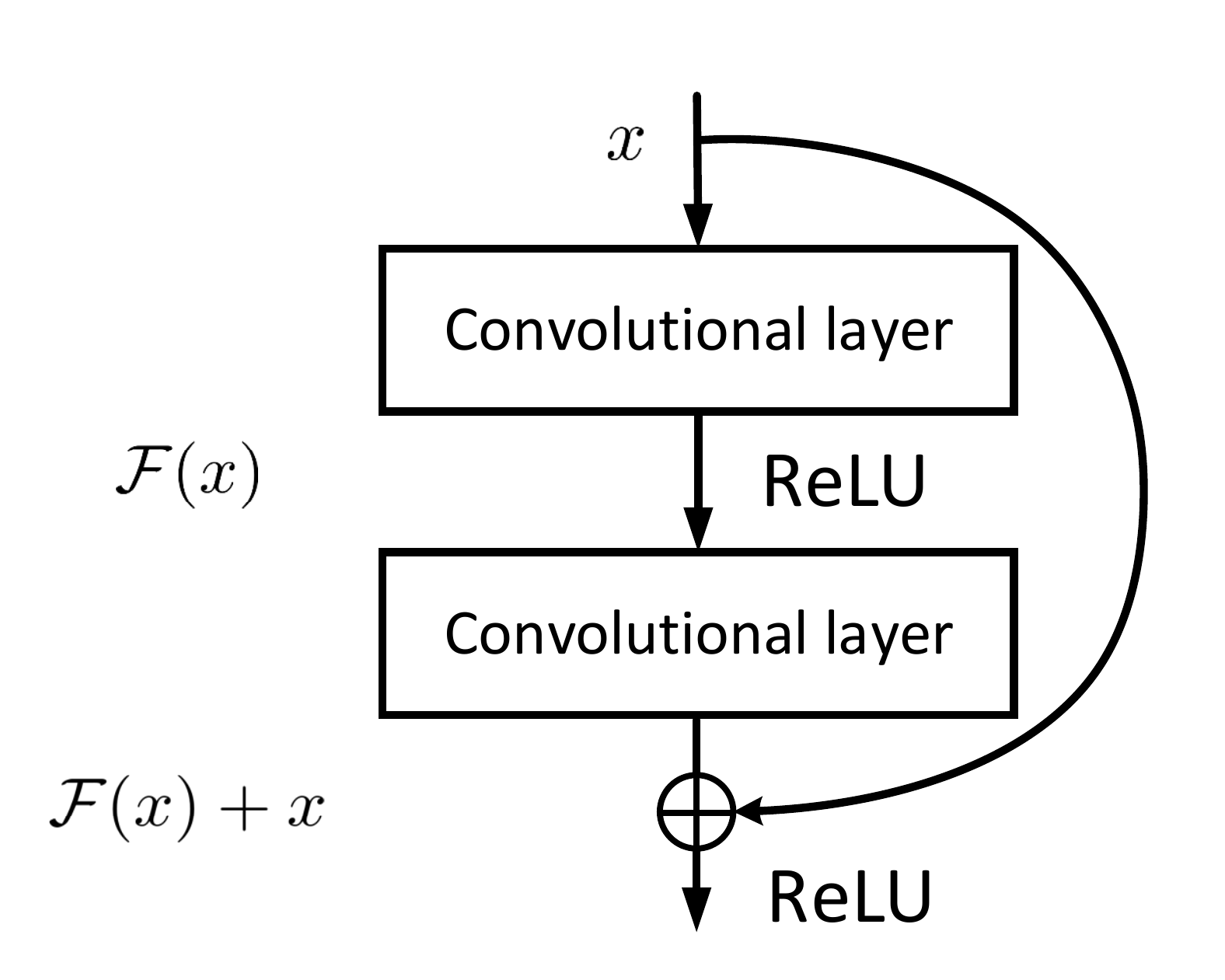}
		
	\end{center}
	
	\caption{The illustrate of identity mapping in ResNet. The figure shows an residual unit including an identity mapping and two convolutional layers.} \label{fig:resnet}
\end{figure}

The general form of the residual unit is defined as
\begin{equation}
\begin{split}
\bm y_l &= \mathcal{F}(\bm x_l) + h(\bm x_l) \\
\bm x_{l+1} &= f(\bm y_l)
\end{split}
\end{equation}
where $h$ is identity mapping and $f$ is ReLU in the original ResNet architecture  \cite{he2015deep}. $\mathcal{F}_{l}$ is composed by a set of Conv units associated with batch normalization, activations and optionally Dropout.
As can be observed, $x_l$ is not completely identity mapping but projected by ReLU after addition.
A follow-up paper \cite{he2016identity} proposes `pre-activation' to allow complete gradient backpropogation during training (explained as follows). `pre-activation' makes the ReLU inside $\mathcal{F}$, which results in the new residual unit definition:
\begin{equation}
\bm x_{l+1} = \mathcal{F}(\bm x_l) + \bm x_l
\end{equation}
This new modification actually is a simple solution to the long-term gradient vanishing issue in CNN training. Specifically, in $l$-th of $L$ residual units of ResNet, the forward output $y_{l}$ and the gradient of the loss $\mathcal{L}$ w.r.t $y_{l}$ is defined as
\begin{equation}
\bm y_{l} = \mathcal{F}_{l}(\bm y_{l-1}) +\bm y_{l-1}
\end{equation}
\begin{equation}
\frac{\partial \mathcal{L}}{\partial \bm y_l} = \frac{\partial \mathcal{L}}{\partial \bm y_{L}} (1+ \frac{\partial}{\partial \bm y_l}\sum_{i=l}^{L-1} \mathcal{F}(\bm y_i))
\end{equation}
Thanks to the addition scheme, the information from prior layers (i.e., $\bm y_{l-1}$ in forward and $\frac{\partial \mathcal{L}}{\partial \bm y_{L}}$ in backward) can flow directly to previous layers without passing to any weight layer. Since the weight layers can vanish the gradient, this property is able to deal with the gradient vanishing effects when training the depth of the network increases \cite{pascanu2013difficulty,he2016identity}.

This simple solution provides subsequent advantages. Several follow-up studies \cite{he2016identity,veit2016residual,huang2016deep} gradually reveal them as discussed in the following.
An remarkable paper worth to mention is stochastic depth network \cite{huang2016deep} built on ResNet. Since training deep network suffers from overfitting and very deep network is usually difficult and inefficient to train. Stochastic depth network trains network with random depth during training stage. Since in each residual network the data from bottom has two paths: $\mathcal{F}(\bm x)$ and $\bm x$, if the data does not pass some $\mathcal{F}(\bm x)$, the actual network depth decreases by some ratios. So during the training, the idea is to randomly block some $\mathcal{F}(\bm x)$ in every mini-batch forward with probability $p$ (i.e. output zeros). Each residual unit could have individual $p$ (named survival rate). The method has a comprehensive discussion of the settings of $p$. While during testing, all $\mathcal{F}(\bm x)$ functions are applied (scaled by $1-p$). Intuitively, this design braces the idea of Dropout and ensemble learning, which significantly improves the generalization of ResNet.

\cite{veit2016residual} argues that ResNet is actually the ensembles of exponential number of relative shallower networks (also mentioned by \cite{huang2016deep}). As mentioned in the last paragraph, each residual unit has two paths to allow data flow to next layers. Suppose we have $n$ such residual units, totally there are $2^n$ number of paths, yielding $2^n$ plain networks with different depths. This paper has experimentally verified its argument. Moreover, it has shown the independence between residual units, in other word, removing some residual units does not influence the results substantially. However, removing a single layer from VGG will cause a huge issue. Swapout \cite{singh2016swapout} pushes the ensemble into an extreme, by combining ResNet, the stochastic depth network, and Dropout.	

DenseNet \cite{huang2016densely} replaces the addition of residual unit with concatenation to allow dense connection between layers, which results in better better feature usage. This strategy implicitly uses multi-layer representations to increase the performance.
Wide ResNet \cite{zagoruyko2016wide} introduces a widen factor and shows that increasing width of layers rather than only depth gives rise to better performance, higher training efficiency, and less memory usage. ResNet of ResNet (RoR) adds another level of identity mapping and shows better performance. ResNeXt \cite{xie2016aggregated} introduces a cardinality factor inside ResNet by repeating homogeneous residual transformation inside a residual unit.

\noindent
\textbf{Generalization} The addition operation is effective in practice for general network training. One main reason is because the addition operation of skip-connection can intrinsically ensemble outputs of modules during forward and equally split the gradients to two paths and may merge them later on during backward. This `averaging' behavior can stabilize the gradients. We have observed various kinds of applications that take benefits from this skip-connection mechanism \cite{lu2016knowing,paszke2016enet,kim2015accurate}. Skip-connection encourages the multi-scale feature fusion to prevent small information loss and makes the network training for efficiency due to better gradient backpropagation. Both characteristics are favorable to medical images.
\cite{drozdzal2016importance} specifically discusses the importance of skip-connection in medical image analysis.

\begin{figure}[t]
	\begin{center}
		\includegraphics[width=0.7\textwidth]{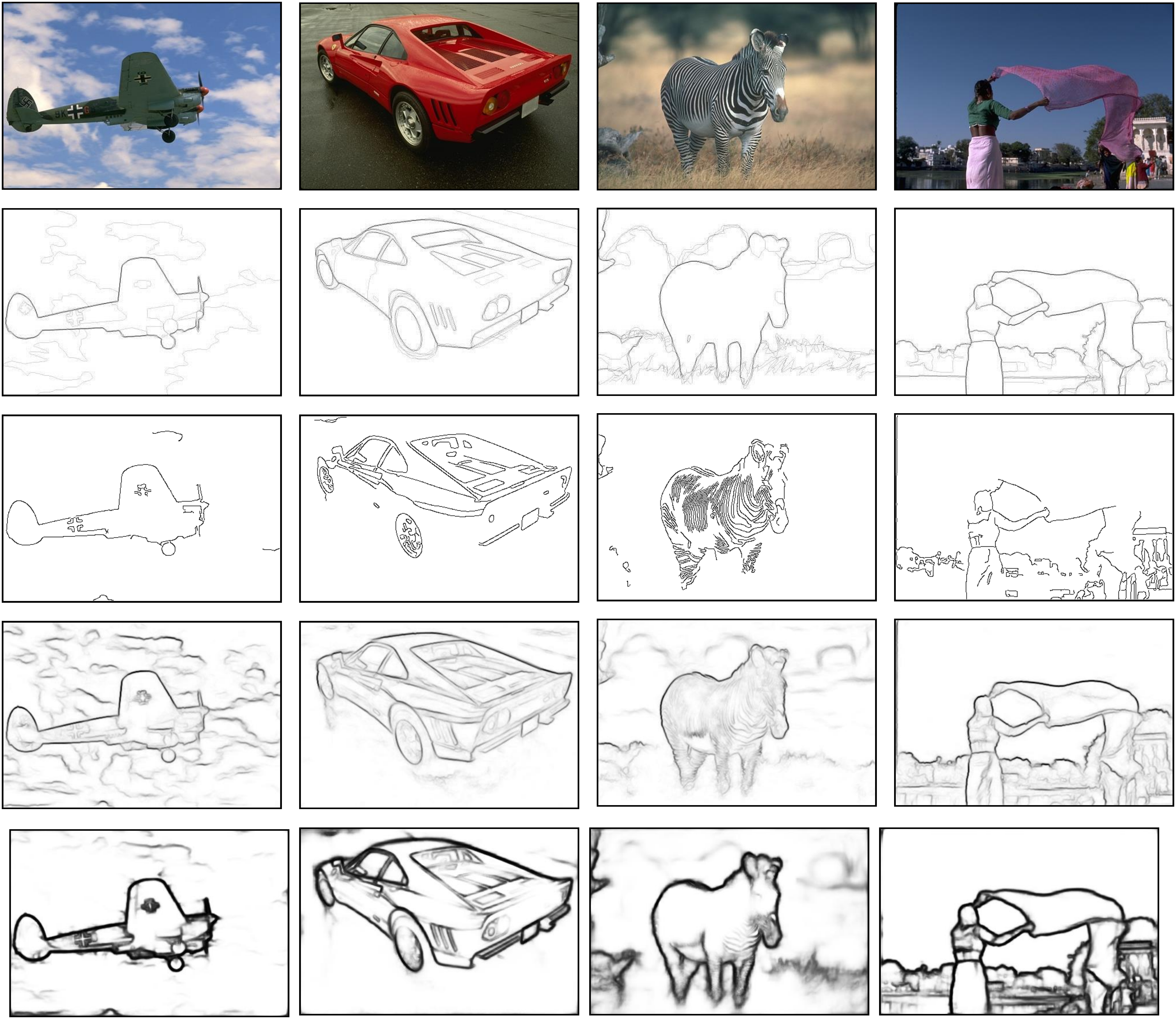}
	\end{center}
	\caption{The visualization of contour detection results on the BSDS500 dataset. The first and second rows show the input image and groundtruth. The third, fourth, fifth rows show the results of basic Canny detector (1989) \cite{canny1986computational}, SE (2013) \cite{dollar2013structured}, and HED (2015) \cite{xie2015holistically} detectors, respectively. SE and HED are learning based methods. As can be observed, current methods can generate more clear edge maps and be aware of object contours and internal or background edges.} \label{fig:edges}
\end{figure}
\begin{table}
	\caption{Comparison of edge detection on the BSDS500 dataset with standard evaluation metrics\cite{martin2004learning}, including F-measure, precision/recall (PR) curves, and average precision (AP). The F-measure score is reported at fixed optimal threshold (ODS) and per-image threshold (OIS). AP is the area under the PR curve. Human annotator accuracy is shown in the first block. The second block shows several early-state unsupervised methods. The third blocks shows conventional supervised methods. The last block shows recent CNN based methods. As can be observed, CNN based methods improve previous approaches by a large margin. } \label{table:edge1}
	
	\begin{center}
		\begin{tabularx}{.9\textwidth}{c|YYY}
			\hline
			~                      & ODS & OIS & AP  \\ \hline
			Human                    & .80 & .80 & -   \\ \hline
			Canny       \cite{canny1986computational}  & .60 & .64 & .58 \\
			Felz-Hutt \cite{felzenszwalb2004efficient} & .61 & .64 & .56 \\
			Normalized Cuts \cite{cour2005spectral}   & .64 & .68 & .48 \\
			Mean Shift  \cite{comaniciu2002mean}    & .64 & .68 & .56 \\
			ISCRA \cite{ren2013image}          & .72 & .75 & .46 \\
			gPb-owt-ucm  \cite{arbelaez2011contour}   & .73 & .76 & .70 \\ \hline
			Sketch Tokens \cite{lim2013sketch}     & .73 & .75 & .78 \\
			SE \cite{dollar2013structured}       & .74 & .76 & .78 \\
			SE-Var \cite{dollar2015pami}        & .75 & .77 & .80 \\
			MCG \cite{arbelaez2014multiscale}      & .75 & .78 & .76 \\
			SE-u \cite{li2015unsupervised}       & .72 & .75 & .76 \\
			PMI	\cite{isola2014crisp}          & .74 & .77 & .78 \\
			SemiContour \cite{zhang2016semicontour}   & .73 & .75 & .78 \\ \hline
			DeepNet \cite{kivinen2014visual}      & .74 & .76 & .76 \\
			$N^4$-field \cite{ganin2014n}        & .75 & .77 & .78 \\
			DeepEdge   \cite{bertasius2014deepedge}   & .75 & .77 & .81 \\
			DeepContour \cite{shen2015deepcontour}   & .76 & .77 & .80 \\
			CSCNN \cite{hwang2015pixel}         & .76 & .78 & .80 \\
			HFL \cite{bertasius2015high}        & .77 & .79 & .80 \\ \hline
			HED \cite{xie2015holistically}       & .79 & .81 & .84 \\
			HED-u\cite{li2015unsupervised}       & .73 & .75 & .76 \\
			CEDN \cite{yang2016object}         & .79 & .80 & .82 \\
			RDS \cite{Liu2016Relaxed}          & .79 & .81 & .82 \\
			COB \cite{maninis2016convolutional}     & .79 & .82 & .85 \\
			PixelNet \cite{BansalChen16}        & .80 & .81 & .83 \\ 
			RCF	\cite{liu2016richer}				&.81		& .83		&	- \\ \hline
		\end{tabularx}
	\end{center}
\end{table}

\section{Image Contour Detection}
\label{sec:edgedet}
In this section, we first briefly review the history literature of edge/contour detection. Then, we introduce the pioneer work of CNN based contour detection methods. Next, we introduce an important end-to-end CNN, supporting present state-of-the-art contour detection methods, and the details of some other breakthrough contour detection methods. Finally, we highlight the shortcomings of existing methods. Figure \ref{fig:edges} qualitatively compares the contour detection results of several strongly foundational contour detection methods on the image contour detection and segmentation benchmark (Berkeley segmentation dataset), BSDS500 \cite{arbelaez2011contour}. Better edge detection methods have stronger ability to detect object contours while ignoring internal edges inside objects or background. Some advanced methods are widely used in medical images but some are barely used. We discuss the strengthens and weakness and clarity their favors to medical images.

In Table \ref{table:edge1}, we compare the contour detection performance of several remarkable contour detection methods under standard evaluation metrics.  We discuss the compared methods in the following.

\subsection{Historical overview of contour detection}

There is a very long and rich history of literature for edge detection \cite{canny1986computational,konishi2003statistical,perona1990scale,marr1980theory,malik2001contour,papari2011edge,shrivakshan2012comparison,ziou1998edge}. The early-stage Canny detector \cite{canny1986computational} computes the local brightness discontinuities and produces continuous edges. Estimating the local changes using global and local cues with multiscales is critical of many following edge detection methods. There are a variety of directions for contour detection using local oriented filters \cite{morrone1987feature,perona1990detecting,freeman1991design}, spectral clustering \cite{cour2005spectral,taylor2013towards,arbelaez2011contour,maire2009contour},  sparse reconstruction \cite{mairal2008discriminative,xiaofeng2012discriminatively}, supervised learning \cite{dollar2006supervised,mairal2008discriminative}, and so on \cite{ma2000edgeflow,comaniciu2002mean,meer2001edge}. We refer readers to \cite{papari2011edge,maire2009contour} for detailed categorization.

There are several remarkable edge detection methods still used by recent state of the arts.
Global probability of boundary (gPb) detector \cite{arbelaez2011contour,maire2009contour}, which is developed by Arbelaez and Malik, computes local orientated gradient features (brightness, color, and texture) based on \cite{martin2004learning} and the multiscaling strategy \cite{ren2008multi}, and then computes global spectral partitions with normalized cut to achieve the globally oriented contour information \cite{maire2008using}. This globalization mechanism differs from the earlier Pb detector \cite{martin2004learning}. This method produces quite clear and effective edges than previous methods. Various methods based on gPb is developed \cite{xiaofeng2012discriminatively,maire2014reconstructive,arbelaez2014multiscale,kim2013learning,yu2015piecewise} for contour detection and segmentation with the focus on both efficiency and accuracy. More importantly, \cite{arbelaez2011contour} also presents an edge based segmentation method called oriented watershed and ultrametric contour map (OWT-UCM) (first published in \cite{arbelaez2009contours}). The overall method for both contour detection and segmentation is well recognized as gPb-owt-ucm in following literature. The OWT-UCM technique and its multi-scaling improved version, Multiscale combinatorial grouping (MCG), are still used by latest CNN based contour detection method to generate object proposals or computing thin edges.
However, gPb is very inefficient for practical usage. After that, the study of edges tends to learning based \cite{dollar2006supervised}, which offers much efficiency and accuracy gains. A critical method is the Structured Edge (SE) detector developed by \cite{dollar2013structured,dollar2015pami}, which is an excellent improvement of SketchToken \cite{lim2013sketch}. SE can be recognized as the most successfully edge detector using random forest with structured outputs \cite{kontschieder2011structured}. The main idea is to use structured learning (i.e.structured random forests \cite{kontschieder2011structured}) to densely predict inherent structures of patches, such as straight lines, parallel lines, curves, T-junctions, Y-junctions, etc. The patch with inherent edge structures is called sketch token in the community \cite{ren2006figure,lim2013sketch}. 
SE is very efficient (60 FPS) compared with gPb-owt-ucm 1/240 FPS and leads the performance of the contour detection field for quite a while, before it is largely surpassed by the introduction of CNN. Many variants of SRF are proposed after then for image edge detection and segmentation \cite{myers2015affordance,arbelaez2014multiscale,uijlings2015situational,teo2015fast,shen2015deepcontour,dollar2015pami,zhang2016semicontour}, and it is also popular in medical images \cite{fujun2015maccai}.

\begin{figure*}[t]
	\begin{center}
		\includegraphics[width=0.99\textwidth]{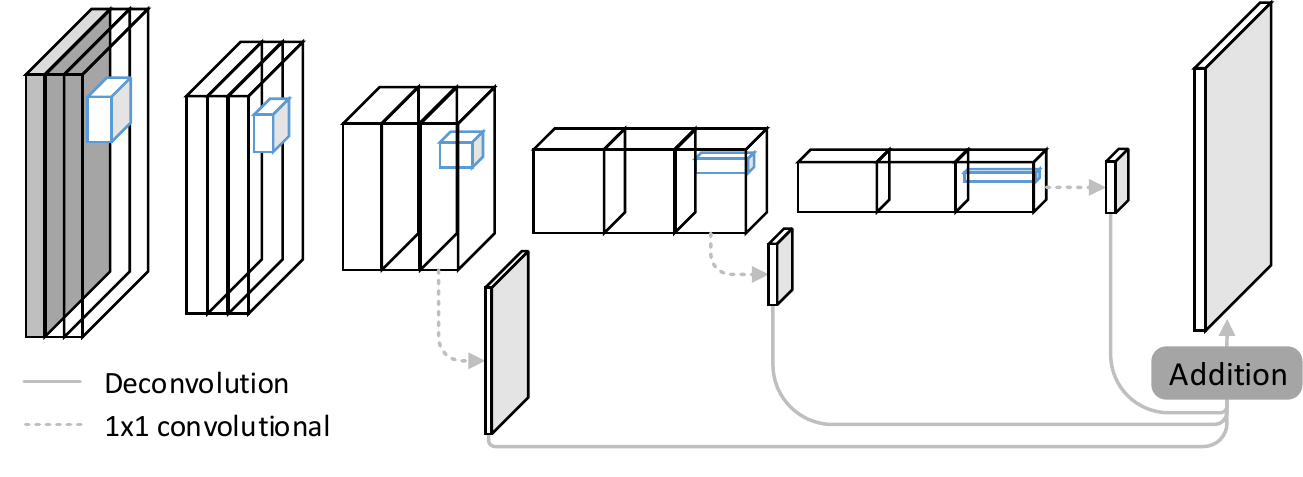}
		
	\end{center}
	\caption{The architecture of FCN. The main part of the CNN architecture adopts the VGG architecture. FCN uses side outputs to extract intermediate layer feature maps. The $1\times1$ convolution is used to generate multi-class segmentation masks. All predictions are unsampled to the image space and added as the final result.} \label{fig:fcn}
\end{figure*}

\subsection{Pioneer CNN-based contour detection}
The performance of edge detection increases significantly in recent two years. We outline the discussed CNN based methods in this section as pioneer work because these methods mostly adopt conventional CNN with the inspirations from previous edge detection methods to build the conception of edge patterns and structures, multi-scaling, and so on.
Following the convention neutral network, the CNN based edge/contour detection takes a local patch around a centering pixel as input and predicts label indicting whether the centering pixel is edge or non-edge. 

\cite{schulz2012learning} propose multiple networks' outputs to perform segmentation and edges.
\cite{hwang2015pixel} (denoted as CSCNN) use  a CNN as feature extractors and train a SVM to classify edges.
\cite{ganin2014n} propose N$^4$ field, i.e., using neural networks and nearest neighbor search to retrieve the best matching edge pattern of local patches.
\cite{kivinen2014visual}, instead of using CNNs, extracts feature using unsupervised generative models (RBM and DBN) and train classifiers to predict edges.
\cite{bertasius2015deepedge} propose DeepEdge, which use multi-scale features from multiple layers and separate two task branches with two losses, one is called classification branch which learns to predict the edge likelihood (with a classification objective) whereas the
other regression branch is trained to learn the fraction of human labelers agreeing about the edge presence at a given pixel. The outputs are combined to predict the edges.

Later on, \cite{shen2015deepcontour} propose DeepContour  to extract visual feature of local patches and use the extracted features as additional features to the used features by the SE detector. The local patches can be categorized into a limited number edge patterns (tokens). Accurately recognizing these patterns is critical for better contour detection. DeepContour uses this property to supervise CNN training to generate rich and discriminative features, then it trains structured random forests to predict edges. This step can be viewed as an edge refinement process. DeepContour achieved leading performance than previous methods.

The efficiency of contour detection is priority. The above methods still suffer from the heavy dense prediction computations, making the edge prediction particularly slower than the SE detector.

\subsection{Fully convolutional network (FCN)}
The introduction of FCN \cite{long2015fully,long2016fully} changes the standard of using CNN for (pixel-wise) dense prediction, making the real-time prediction becomes possible and as well largely improved performance, hence it benefits later contour detection methods. 

The technique of FCN is actually straightforward and simple. FCN allows the network directly outputs a segmentation mask having the same dimension of the input. Suppose the input RGB image has dimension $3\times H\times W$. The output of image will have the size $C \times H \times W$ for $C$ class semantic segmentation. In other word, each spatial location of the segmentation mask predicts the probability of its semantic label. Figure \ref{fig:fcn} illustrates the FCN architecture.

FCN uses VGG as the basic network architecture. We have discussed the details of VGG previously. It contains 5 convolutional sets, with 5 $2\times 2$ max pooling in between, which totally resizes the original input by $2^{5}$. So the feature map dimension of the last convolutional set is $512 \times \lfloor \frac{H}{32}\rfloor \times \lfloor \frac{W}{32}\rfloor$.  These feature maps keep coarse spatial location where each spatial location stands for $32 \times 32$ region in the original image.

Instead of using the fully connected layer after the last convolution layer as CNNs for image classification, FCN directly applies an $1\times1$ convolutional to transform the $512$-dimensional vector of each spatial location to the label space, i.e., $C$-dimensional in this example, as the semantic label probability distribution. Next, FCN adds an deconvolution (used as an upsampling operation) layer to enlarge feature maps from $C \times \lfloor \frac{H}{32}\rfloor \times \lfloor \frac{W}{32}\rfloor$ to  $C \times H \times W$, resulting a pixel-wise segmentation mask. It is worth to mention that the terminology of deconvolution used here is debatable, because it is not the conventional deconvolution operation \cite{zeiler2010deconvolutional}. The one used here is actually a 'backward convolution' operation, i.e., each spatial pixel performs element-wise product with all the weights of the kernel and expand the predictions (one for each weight value) as an image extent. Figure \ref{fig:unpool} explains this `deconvolution' operations. The weights of the deconvolutional layer can be initialized as a bilinear interpolation kernel and allow this deconvolution to act upsampling behavior. We name it upsampling in this paper because we will introduce another strategy (i.e, unpooling plus deconvolution) next to achieve structured outputs (see bottom of Figure \ref{fig:unpool} ). Training the network is straightforward. In the loss layer, every pixel contributes to the loss, all spatial locations are summed. This variation does not break direct backpropagation. This training strategy, i.e., inputting an image and outputting a pixel-wise prediction map, is termed as end-to-end training.

This simple approach achieves surprising good results compared with previous methods and the prediction is very efficient (less 1 second for an $500\times 500$ image) because the network does not need computational expensive fully connected layers and only once forward is needed to obtain the final results. To generate more precise and robust prediction, FCN also proposes to build side outputs to make use of the feature maps from multiple convolutional layers. Since convolutional layers have different feature map dimensions, this approach in nature utilizes multi-scale information. FCN-32s uses the one side-output to predict the segmentation. FCN-8s uses three side-outputs and merge as the final results as shown in Figure \ref{fig:fcn}.

This idea becomes the foundation of CNN based image segmentation and contour detection crossing various image domains. Wide improvements have been proposed in recent years \cite{ pinheiro2015learning, papandreou2015weakly,zheng2015conditional,lin2015efficient,lin2015deeply,hariharan2015hypercolumns,sharma2015deep,chen2014semantic}

\begin{figure}[t]
	\begin{center}
		\includegraphics[width=0.7\textwidth]{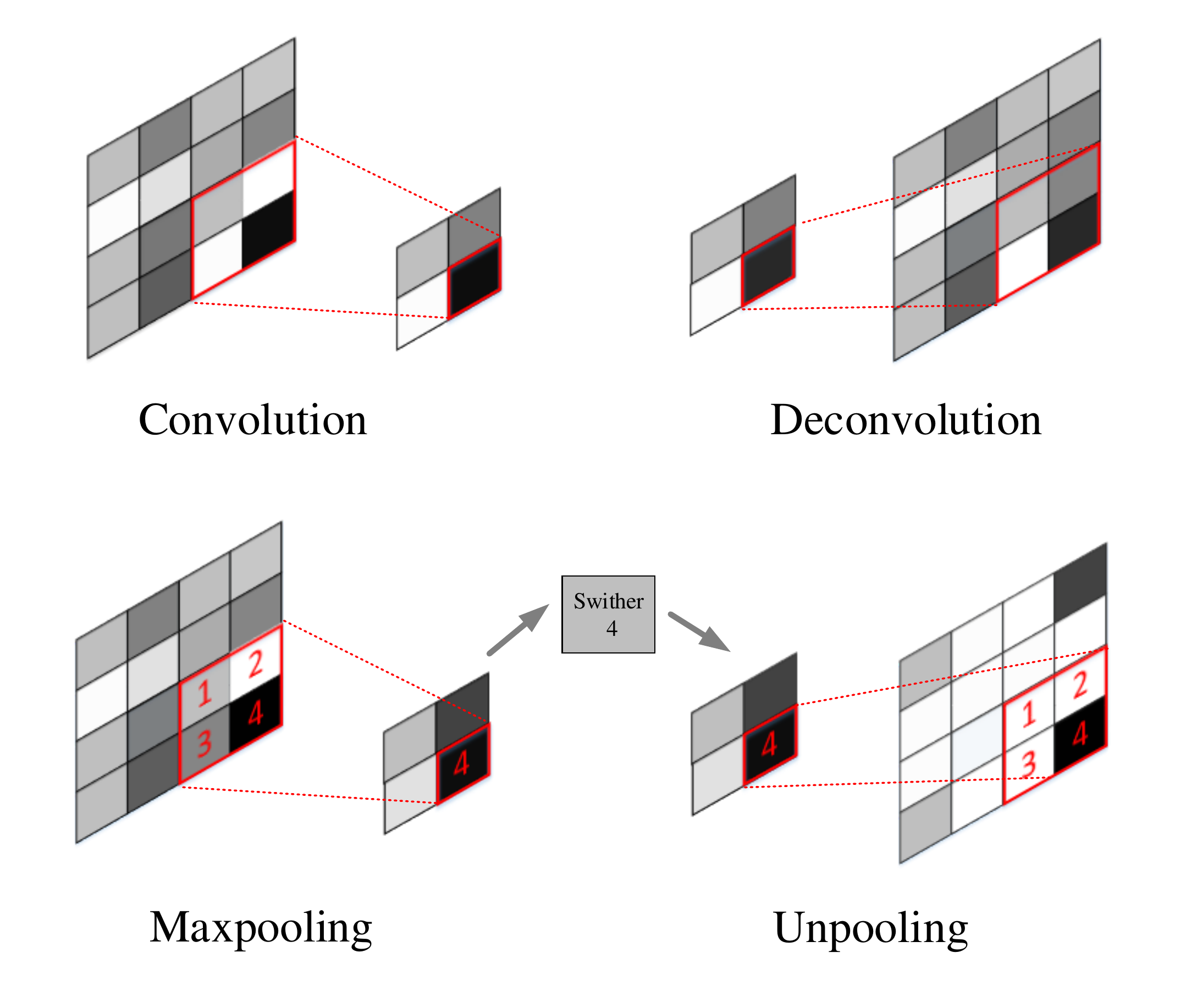}
		
	\end{center}
	
	\caption{The illustrate of convolution and deconvolution operation and maxpooling and unpooling layer. The kernel size is $2\times2$. Convolution operation averages the values of a $2\times2$ region. Deconvolution computes the values of the region value given one value. Maxpooling selects the maximum response in each region. Unpooling uses the switcher information (saved the selected location of the region in the maxpooling layer) to fill a value in the region, yielding a sparse feature map. } \label{fig:unpool}
\end{figure}
\subsection{Holistically-nested edge (HED)}
HED is the first method successfully applied FCN into contour detection. It significantly improved previous methods. The main contribution of this paper is the usage of FCN with side outputs and deep supervision \cite{lee2015deeply}. Actually, the concept of side output uses extra branches from intermediate layers to encourage multi-scale feature reuse. 
Differently, the side outputs of HED are not directly combined together to output a single mask (edge map here) as FCN
does or discrete label as DeepEdge does. Each side output is directly used to predict the edge map. Since side-outputs connecting intermediate layers have different feature map dimensions, deconvolution layer (upsampling layer) is used to resize the feature map to the original input image size. Each mapped output is passed to a Sigmoid cross-entropy loss to perform pixel-wise binary prediction. The loss of the network is define as
\begin{equation}
\mathcal{L}_{side}(\bm W, \bm w) = \sum_{i=1}^{S}  \alpha_i \mathcal{L}^{i}_{side}(\bm W, \bm w^{(i)}),
\end{equation}
where $ \mathcal{L}^{i}_{side}$ is the loss from $i$-th side output. $\bm W$ is the core network parameters. $\alpha_i$ is the loss weight. $S$ is the number of slide outputs. Sine HED uses VGG16 as the base network architecture. $\bm W$ refers to the parameters of its $5$ convolutional units. $\bm w^{(i)}$ is the parameters of $i$-th side output. It is actually a $1\times1$ convolutional to map the feature map to the label space (edge or non-edge).

Since edge pixel has much small portion than non-edge pixel in one image, the loss is imbalanced per image. From machine learning perspective, unbalanced training data is not undesirable for model optimization.
HED introduces weight-balanced loss, defined as follows:
\begin{equation}
\begin{split}
\mathcal{L}_{side}^{(i)}(\bm W, \bm w) = - \beta \sum_{y \in Y_{+}} log Pr(y_j=1|X; \bm W, \bm w^{(i)} ) \\
- (1-\beta ) \sum_{y \in Y_{-}} log Pr(y_j=0|X; \bm W, \bm w^{(i)} ),
\end{split}
\end{equation}
where $\beta = |Y_{-}|/|Y|$ and $1- \beta = |Y_{+}|/|Y|$. $|Y_{+}|$ denotes the number of pixels belong to edges according to groundtruth edge map $Y$. $\beta$ is the weight balancing coefficient. Besides individual loss, all slide outputs are then fused together and generate a fused output associated with a loss:
\begin{equation}
\mathcal{L}_{fuse} = Distance (Y, \; \sum_i^{S} \gamma_i \hat{Y}^{(i)} ),
\end{equation}
where $\hat{Y}^{(i)}$ is the predicted edge map by $i$-th side output. and $\gamma_i$ is the learnable weights. $Distance$ is also a Sigmoid cross-entropy loss to measure the pixel-wise prediction error.

The training of this method is relatively light. It uses pre-trained VGG-16 as initialization and train with a few thousands iteration to obtain the results. Note that BSDS500 only has 200 training image. HED sample 100 test images and test on the rest 100 images. In addition, the paper also discuss the inconsistence of groundtruth. The outputs at deeper layers are coarse which will in nature ignore detailed edges (such as background and object internal edges). To prevent very fine annotations (containing many `noisy' edges) of groundtruth from affecting the convergence of supervision of deeper layers, the paper treats a pixel as edges only it is labeled as edge in at least three annotations.

Side-output with deep supervision is fairly effective combination to boost the performance of dense prediction tasks, because it maximizes the reuse of rich hierarchical representations at different layers with different scale. The multi-scaling property is also a well-known approach to improve the contour detection accuracy.
Most following work lies on the feature map re-usage to push the performance. For example, \cite{kokkinos2015pushing} also trains multi-scale HED based on multi-scale inputs to further boost BSDS500 accuracy. It has outperformed the empirical accuracy of human annotator (.80 ODS). RCF \cite{liu2016richer} generalizes HED by using richer features from all convolutional layers of a CNN, pushing the performance to .81ODS.

HFL \cite{bertasius2015high} uses object-level features to accomplish low-level edge prediction, because the human vision system uses object-level reasoning to locate edge points. Specifically, it extracts object-level deep features from multiple layers of VGG-16 and use a MLP to classify edges. This method can be viewed as a special way to use rich features from a pre-trained CNN. More interestingly, HFL extends its network to the application of semantic boundary labeling and semantic segmentation to show that low-level boundaries have positive effects to high-level vision tasks.

Another analogous method is PixelNet \cite{BansalChen16}, which highlights the usage of all feature maps and uses a specially designed predictor (a MLP) for coherent semantic segmentation and contour detection. PixelNet also discusses the sampling of predicted pixels and mini-batch to reduce the unbalance of edge and non-edge pixel ratio and the memory consumption.

In addition, HED has been applied and improved for different tasks. \cite{shen2016object} uses HED to extract object skeleton. \cite{li2015unsupervised} even train HED in an unsupervised manner base don video optical flow by iteratively refining the model.  Beyond the natural image domain, HED is widely welcomed in medical image domain because of its efficiency and multi-scaling scheme to handle resolution and scale problems ubiquitous in medical images. We will discuss them in the following section.
There also several notable papers generalize HED \cite{xu2016gland,cai2016unified,teikari2016deep,xu2016gland} for various computer vision tasks in both natural image and medical image domains.
\begin{figure*}[t]
	\begin{center}
		\includegraphics[width=0.99\textwidth]{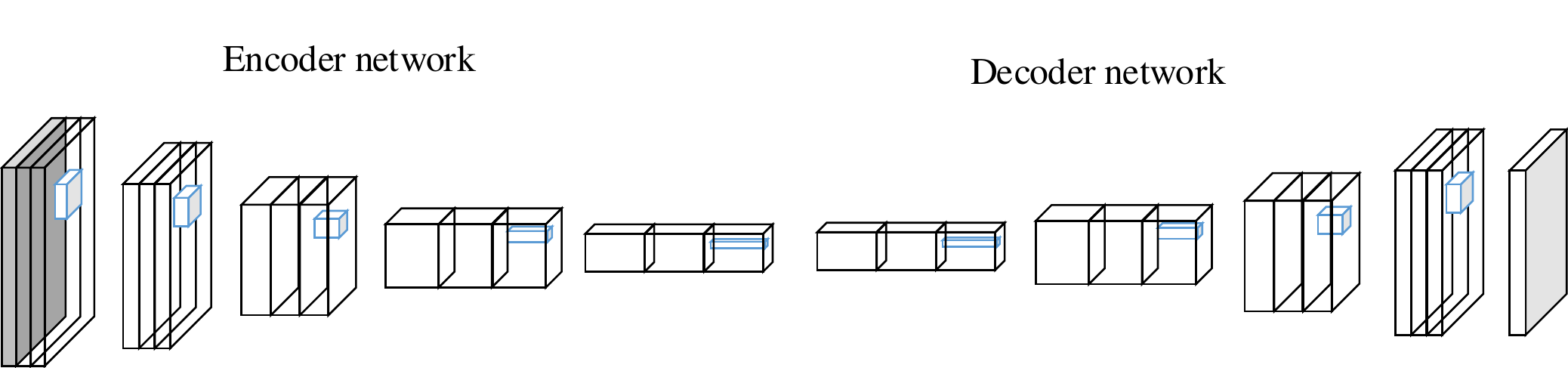}
		
	\end{center}
	
	\caption{The architecture of DeconvNet. DeconvNet includes an encoder to encode an input image to a set of feature maps and the decoder to decode the encoding to a segmentation mask through a set of unpooling layers (illustrated in Figure \ref{fig:unpool}) and convolutional layers.  } \label{fig:deconv}
\end{figure*}
\subsection{Encoder-Decoder network}
Another noticeable network for contour detection is called Encoder-Decoder network \cite{yang2016object}. This network has very similar architecture with DeconvNet \cite{noh2015learning} for semantic segmentation. The difference is that the encoder and decoder  architectures are asymmetric. The encoder part is identical to DeconvNet borrowed from VGG-16, but the decoder part has light computational units. Instead of using deconvolution layer, every unpooling layer is followed by a convolutional operation. Figure \ref{fig:deconv} illustrates the network architecture. 
This network is a successful application of encoder-decoder style CNN architectures on contour detection.

Different from methods using side outputs or deep supervision, this method directly uses the last convolutional layer of VGG-16 (this encoder part is fixed during training) and uses a set of unpooling and convolutional to generate unsampled prediction map. The feature map dimension of the last convolutional layer is $32$ times smaller than the input image. Small scale coarse prediction ignores the detailed information such as short and weak edges, thus  the generated contour map ignores the background and object internal edges but only retains occlusion boundaries \cite{sundberg2011occlusion}. So this method achieves significantly improved performance than HED ($57.0$ ODS vs. $44.0$ ODS) on the PASCAL val2012 dataset. Note that this dataset contains only groundtruth of object contours. However, when applied onto BSDS500 dataset where groundtruth contains fine edge annotations, this method performs slightly worser than HED.

Compared with HED, this network is better capture overall object contours using content information rather than small edges due to gradient changes. This property is suitable to detect contours of medical image objects, such as organs.

\subsection{Oriented contour detection}
Using the global orientation to predict edges has been well studied previously \cite{shi2000normalized,arbelaez2011contour}. Affinity matrices well capture pairwise local intensity changes by using a global graph embedding and the generated graph partitions preserve strong edge information where each map highlights edges with one particular orientation.
There are a few work considers the orientation information inside the CNN training.

Affinity CNN \cite{maire2015affinity} directly trains a network to output the affinity matrix. To achieve that, this method trains 48 predictors (composed by convolutional layers) to predict the affinity of each pixel to its 8 neighbors at 3 different scales (distances of 1, 4, and 16 pixels). There are two losses functions supervised by pre-computed affinity matrix and image edge groundtruth respectively.

Another remarkable work is called Convolutional Oriented Boundaries (COB) \cite{maninis2016convolutional}. Firstly, COB generalizes the side output and deep supervision scheme of HED to obtain fine and coarse contour maps. To estimate the contour orientation, COB connects multiple small sub-networks to predict oriented edge maps for each orientation bin. Each subnetwork is access to all side outputs with different scales. To decide the final orientation of each pixel, COB computes the max response of sub-network outputs respecting to different bins and may average the two orientations if both oriented sub-networks have high responses. COB uses a strong 50-layer ResNet as the basic network while most other methods we discussed use VGG-16. 
COB shows that using ResNet improves the performance by a quite large margin. It demonstrates the important role of the basic CNN for high-level applications.

Using orientation information is useful for medical images. For example, in Lung X-ray image diagnosis, healthy images contain clear rib cage contour \cite{dai17scan}. In this way, we can predict contours with semantic information to help diagnosis.

\subsection{Weakly-supervised, semi-supervised and unsupervised edge detection}
Obtaining the annotated contour detection dataset is very labor expensive. At present, only the BSDS500 dataset has fine edge annotations. Dense prediction tasks can obtain many pixel-wise label from a single image to optimize the parameters of CNNs, however we can still observe that the size of training data is determinate \cite{yang2016object,maninis2016convolutional}. There are some yet limited work that studied how to introduce unlabeled data to train or improve an edge detector \cite{li2015unsupervised,zhang2016semicontour,Liu2016Relaxed,khoreva2016weakly}.
However, we only witnessed growing related literatures for semantic segmentation \cite{hong2015decoupled,hong2015learning,papandreou2015weakly,dai2015boxsup}.

\cite{li2015unsupervised} propose an unsupervised learning method of edges, which utilizes the property of discontinuity around the edge pixels of motions to generate weak annotations from a large video dataset. This method uses a quite sophisticated process to generate clean annotations based on motion estimation techniques including motion estimation, motion contour detection, and edge alignment. Using motion cues for contour detection is proposed by \cite{sundberg2011occlusion}. Then, it treats the motion estimation and contour detection as an iterative optimization process. At each iteration, the estimated edges are used as supervision to train an edge detector. The trained edge detector is used to generate edge maps for better motion estimation for next iteration. The paper conducts experiments using both SE and HED as the base edge detector, showing than this kind of coarse-to-fine interaction training strategy does improve the performance of CNNs and structured random forest classifiers. However, as the author stated, the generated annotations have noises and have unable to generate fine edges, so it can hardly obtain the same or even outperform the edge detector trained with strong supervision with human annotations although more images are available. Besides unsupervised training, \cite{zhang2016semicontour} propose a semi-supervised structured ensemble learning method, which is built on SE, to train an edge detector with only 3 labeled images and outperforms this unsupervised method (see Table \ref{table:edge1}). However, this semi-supervised learning method can not be used for CNNs. There is no work study about semi-supervised CNNs for contour detection. \cite{khoreva2016weakly} generate object contour supervision under multi-level supervision and test the SE and HED detectors. 

The coarse-to-fine CNN training paradigm for contour detection is embedded into intermediate layers of a single CNN by \cite{Liu2016Relaxed}. This supervision is denoted as relaxed deep supervision (RDS), which is used to improve a pre-trained HED with a large set of coarse edge annotations. The motivation behind is that RDS relaxes the human annotations (edge and non-edge points) to get more relaxed labels, which is used to adapt to the diversities of intermediate layers. Relaxing labels is produced by Canny, SE or HED detectors. The benefit of RDS is that it processes the false positives using a ``delayed strategy'' to allow more discriminative layers (deeper layers) handle difficult points and leave these difficult points ignored in early layers. This is a validate way to achieve better network convergence.

Using less annotated data for training CNN is obviously essential for medical image analysis.
Unfortunately, this area is waiting to be explored.

\begin{sidewaystable}
	
	\caption{Overview of the literature using deep learning for various kinds of medical image segmentation that are contour aware. Most methods use end-to-end CNNs (FCN or HED) scheme to perform pixel-wise classification. See text for more detailed explanations.} \label{tab:medseg}
	\begin{center}
		\footnotesize 
		\begin{tabularx}{\textwidth}{ l|l|X } 
			\hline
			Reference  &  	 Task &						Method  \\ \hline
			\cite{ciresan2012deep} 	& Neuronal membrane segmentation	 & 	 Standard CNNs with patch-wise pixel classification. 	\\
			\cite{ganin2014n}	& Retinal vessel segmentation & Using CNN features to model edge patterns and apply nearest search to detect edges.		\\
			\cite{ronneberger2015u}	& Biomedical image segmentation  &   A novel end-to-end CNN (Unet) for cell segmentation with special designs for contour preserving.																																					\\
			\cite{SoZh15}	&	Cervical cell segmentation		&	Extracting CNN and different kinds of features to learn to localize object contours.		\\
			\cite{su2015robust}		& Cell segmentation	&    An autoencoder based method to recover broken edges between touching or overlapping cells. \\
			\cite{chen2016dcan}	& Gland segmentation  &    An end-to-end CNN uses multi-layer feature maps to perform contour detection and segmentation. 		\\
			\cite{chen2016deep}		&	Neuronal membrane segmentation & 	  An deeply supervised CNN to capture contextual information.																												\\
			\cite{maninis2016deep} & Retinal image segmentation	& Using mixed multi-layer feature maps to segment vessels and optic disc. \\
			\cite{xu2016gland}		&	Gland segmentation	&   Combining FCN and HED to conduct segmentation and edge detection together.	\\
			\cite{roth2016spatial}	& Pancreas segmentation		& 	Spatial aggregation with random forest to combine edge and interior CNN cues of organs.																								\\
			\cite{Cai2016} 		&	Pancreas segmentation				& 	Aggregating organ contour detection and segmentation results of  HED and FCN through conditional random fields. \\
			
			\cite{dou20163d}	& Liver segmentation	&  A 3D deep supervised FCN with an extra contour refinement process.		\\
			\cite{chen2016combining}& Biomedical image segmentation	&  Using RNNs to model the intra-slice and inter-slice context represented by FCN to separate touching objects. \\
			\cite{chen2016deep}	&				& \\
			\cite{cciccek20163d} &	Volumetric Segmentation	& The 3D version of Unet.		\\
			\cite{milletari2016v}	&	Volumetric Segmentation			&	VNet with new dice score driven loss function to trade-off imbalance of labels.			\\
			\cite{rupprecht2016deep}& MRI segmentation	 & 	Using CNNs to predict evolution of active contours for contour localization.		\\
			\cite{milletari2017hough}&  Ultrasound and MRI segmentation	& A hough-voting CNN has robust contour extraction of anatomy.	\\
			\cite{yangshape}		&	Ultrasound segmentation	&	Using recurrent memory networks for contour completion.															\\
			\cite{MoLi17} & Cardiac MRI	& 	Combing a dynamical system and a CNN to model the contours of objects.		\\
			\cite{oktay2017anatomically} & Multi-modal cardiac & Incorporating prior anatomical knowledge (e.g. boundaries and shape) into network training.\\
			\hline
		\end{tabularx}
	\end{center}
\end{sidewaystable}
	
\section{Medical Image Contour Detection and Segmentation with CNNs}
\label{sec:meidcaledgedet}
Deep learning has became the mainstream of medical image contour detection and segmentation methods \cite{jiang2010medical,xing2016robust,greenspan2016guest,shin2016deep,roth2016improving,bentaieb2016topology,oktay2017anatomically,cai2017improving}.
In this section, we review recent CNN based methods for medical image contour detection and segmentation. Most reviewed paper aims at segmentation. We focus on the methods that use the abovementioned methods or are designed to be contour aware to achieve more accurate segmentation. Table \ref{tab:medseg} summarizes the reviewed methods. Detailed are discussed in the following.

\cite{ciresan2012deep} adopt the standard CNN as a patch-wise pixel classifier to segment the neuronal membranes (EM) of electron microscopy images. This study is a pioneer work of using CNN for medical image segmentation. It won for ISBI 2012 EM image segmentation challenge and significantly outperforms other competing methods. In \cite{FaPe16}, a CNN with an architecture specifically optimized for EM image segmentation is presented. Compared to the original CNN, smaller receptive field in the upper layers and deeper architecture are employed. Such optimized design enables the network to learn better features from the local context with increased non-linearity. Significant improvement in performance is validated on the ISBI 2012 challenge \cite{arganda2015crowdsourcing}. 
\cite{SoZh15} propose a segmentation system for cervical cytoplasm and nuclei in which pixel-wise classification is obtained by multiple convolutional networks trained for images at different scales. Then they use various kinds of features to learn to localize object contours and split the touching objects.

The majority of contour detection and segmentation methods follows the structure of the FCN \cite{long2015fully} and HED \cite{xie2015holistically} networks. \cite{ronneberger2015u} propose U-net, an end-to-end CNN that can take advantage of information from different layers. To handle touching objects, a weighted loss is introduced to penalize the errors around the boundary margin between objects. The proposed U-net achieved the best performance on ISBI 2012 EM challenge dataset \cite{arganda2015crowdsourcing}. The state-of-the-art segmentation performance on the EM dataset is achieved by a new deep contextual network proposed in \cite{chen2016deep}. The deep contextual network adopts an architecture that is similar to HED. The difference is that the final segmentation result is a combined version of the segmentation results derived from different layers through an auxiliary classification layer. In the forward propagation, such design can more efficiently exploit the contextual information from different layers for edge detection. In return, the lower layers can be deeply supervised through these auxiliary classification layers. This is because the classification layers provide a short cut between the lower layers and final segmentation error. \cite{chen2016dcan,chen2017dcan} propose a deep contour-aware network for gland image segmentation. This method uses side outputs as multi-tasking deep supervision. The detected contour map is merged with the segmented binary mask to prevent touching of glands, which is a special treatment to cell contours. This method won the 2015 MICCAI Gland Segmentation Challenge \cite{sirinukunwattana2016gland}. In the following, \cite{xu2016gland} propose a multichannel side supervision CNN for gland segmentation. This network can be treated as a combination of HED and FCN for simultaneous segmentation and contour detection. Similarly, \cite{nogues2016automatic} propose a lymph node cluster segmentation algorithm based on HED, FCN and structured optimization to address the contour appearances. \cite{Cai2016} propose a data fusion step using CRF to adaptively consider the segmentation mask generated by FCN and the contour map generated by HED for pancreas segmentation. \cite{roth2016spatial} propose to use random forest based spatial aggregation to integrate semantic mid-level cues of deeply-learned organ interior and boundary maps to segment pancreas with HED.\cite{maninis2016deep} explores the combination of multi-layers' feature maps to perform multi-task learning on vessel segmentation of retinal images. \cite{oktay2017anatomically} proposes to incorporate anatomical priors on anatomy (e.g. shape and label) structure into CNNs, so as to make the predictions anatomically meaningful, especially for the case when input images have missing boundaries. 

CNN based methods for 3D medical image segmentation have been attracting attentions in recent two years. Most existing methods are extensions of known 2D CNNs. \cite{dou20163d} propose a 3D deeply supervised network for Liver segmentation. It can be viewed as a 3D extension of HED. Moreover, it uses a fully connected CRF to refine the object contours. 
3D Unet \cite{cciccek20163d} is proposed by the same group with U-net for 3D volumetric segmentation.  \cite{milletari2016v} propose V-Net, which contains a new loss function based on Dice coefficient to resolve the strong imbalance between foreground and background. It uses the skip-connection strategy of Unet to prevent the detail information loss which will affect fine contour prediction.

Besides the direct application of end-to-end CNNs for pixel-wise classification, there are a number of interesting studies exploring the usage of CNNs or RNNs to achieve better context information modeling (e.g. contour completion). \cite{su2015robust} propose to use stacked denoising autoencoder to restore the broken cell boundaries for cellular segmentation in brain tumor and lung cancer pathology images.  Similar to \cite{su2015robust}, \cite{KaPe16} propose a breast density segmentation method based on a multi-scale CNN that is trained in an unsupervised way in which an autoencoder is trained. During the unsupervised training, image patches and their corresponding segmentation masks are randomly cropped and the network is trained to reconstruct the segmentation masks. The unsupervisedly learned model is used to extract features for pixel level classification. Therefore, the different components are delineated. \cite{yangshape} use RNNs to achieve completion for ultrasound images where the contours are unclear and broken. The RNN it used is called bidirectional Long Short-Term Memory (BiLSTM) networks, which is able to leverage past and future information to make prediction. \cite{chen2016combining} use RNN model and propagate the contextual information of the third dimension of the 2D image planes. A 2D CNN (i.e. U-net) extracts the hierarchy of contexts from 2D images and pass the information to RNN to leverage the inter-slice correlation for 3D segmentation. Their good results on fungus images, which contain very weak boundaries between objects, demonstrate its ability of be aware of object contours. 
\cite{rupprecht2016deep} trains class-specific convolutional neural network to predict the evolution of active contours as a vector field. This method is a new way to formulate the structured output of CNNs. \cite{milletari2017hough} propose hough-CNN, a CNN with voting mechanism along the contours of objects to localize anatomy centroid.  A recent work \cite{MoLi17} combines CNNs with dynamic system theory for Cardiac organ contour detection. This method takes advantage of an important concept
in dynamical system, i.e., limit cycle, to represent the contours of the target object. Instead of classifying pixels into label classes, they propose to predict a vector for each pixel and thus a vector field is formed for an entire image. Based on the vector field, the organ contour is detected through dynamic theory in which a limit cycle is detected as the finally detected contour. The method needs very limited training data to train the model.
	
\section{Discussion}
\label{sec:discussion}
We have discussed the state-of-the-art image edge or contour detection methods in the computer vision community and we review their applications in the medical image domain. 
Based on the discussed methods above, it can be observed that the usage of CNNs in medical image contour detection and segmentation is relatively crowded into a narrow line. Most work leverages on end-to-tend CNNs for direct dense prediction. Extension towards to wide and new perspectives to solve specific problems would be necessary for CNN development in medical image analysis, but only a few literature exists. We discuss some interesting topics and outline potential directions.

\textit{\textit 1) Multi-scaling} Medical images intrinsically contain rich multi-scale information such as the nucleoli and tumor regions in microscopic images. Using fine features without much spatial information loss is important for contour detection. From the popularity of HED in medical images, sufficient usage of the layers' feature maps is as always promising. 
Moreover, the aggregation of HED is a way of model resembling \cite{kokkinos2015pushing}. Ensembling offers a multi-scaling and averaging mechanism, which is important to generate smooth contours. Detailed explorations of state-of-the-art CNNs, for example, ResNet \cite{he2015deep} and DenseNet \cite{huang2016densely}, are necessary, which have skip-connection to strengthen feature map usage.  Appropriate usage of skip-connection to build very deep FCN is studied by \cite{drozdzal2016importance}.
RCF \cite{liu2016richer} shows an extreme of using features from convolutional layers. Most studies enable multi-scaling with side outputs and optional deep supervision. From our experience, the supervision at shallow layers usually has large losses which is very difficult to overfit even on training data. Large losses will result in large gradients and thereby disturb the error backpropagation of deeper layers. This is also discussed by \cite{Liu2016Relaxed}. We think there should be more careful studies on the consideration of effective deep supervision mechanism.

\textit{2) Transfer learning} Transfer learning is gaining popularity in the medical imaging domain. Several literature \cite{tajbakhsh2016convolutional,shin2016deep,greenspan2016guest} have shown that fine-tuning CNNs trained on natural image datasets helps improve the performance. Designing highly effective network architectures needs rich experience, but borrowing or modifying existing architectures alleviates the pains. Transfer learning addresses the insufficient dataset problem in the medical domain. In fact, dense prediction tasks with end-to-end CNNs can implicitly gather many training data (one for each pixel). That is one of the reasons that some large architectures like U-net \cite{ronneberger2015u} can be trained from scratch using only 30 images. However, we believe transfer learning will give further improvement, with careful consideration of the usage of shallower layers and deeper layers. Shallower layers capture fine edge information which are shared between natural images and medical image, while the deeper layers capture content information which are completely difficult. Therefore, the appropriate usage of feature maps of earlier layers is important (which also recalls the problem of multi-scaling), for example, extra links to combine shallower layers and deeper layers \cite{milletari2016v}.

\textit{3) Discontinuity of broken edges} Although the CNN is powerful to detect edges. The severe and common touching and overlapping phenomenon between objects in medical images are still challenging. One common way is to deploy a remedy process on CNN outputs as discussed above, by integrating conventional methods \cite{fujun2015maccai} or extra deep learning models attempting to reconstruct a better contour map \cite{su2015robust}. 
RNN has shown the potential to achieve contour completion \cite{yangshape} with its ability to capture long-term dependence of inputs. However, the semantic information of edges is very limited. To enrich the features of edge as the input of RNN, better usage of feature maps would be helpful. In addition, instead of conventional RNNs, advanced memory networks \cite{sukhbaatar2015end,graves2016hybrid} could be useful to handle contour reasoning.
In addition, using shape priors \cite{xing2016transfer} is popular to main the structure of objects in medical image segmentation because organs or cells usually have similar shapes. \cite{chen2013deep, eslami2014shape} use deep Boltzmann machines to model hierarchical structures to constrain the evolution of the shape-driven variational models. \cite{7351118} considers using CNNs to achieve a similar task. However, we haven't seen studied to incorporate shape priors into CNN training. We think there are several direction can be considered. The first is adding shape prior constraints in the final loss layer. The second is formulating other structured outputs to maintain the shape and continuity of predicted contours, such as \cite{rupprecht2016deep,MoLi17}.

\textit{4) Miscellaneous} Collecting large-scale medical image dataset is extremely difficult. Exploring using less-labeled training data is essential. 
\cite{khoreva2016weakly,li2015unsupervised} have show ways of using CNN  on natural images and show promising results. 
However, the study on medical images is not seen. Low-quality or high SNR images are also common in medical images. Specific methods to resist such situation is necessary \cite{ofir2015fast}.

\section{Conclusion}
\label{sec:conclusion}
This paper discusses the key components and technical ingredients of CNNs specific to medical image contour detection. 
Specifically, we review several mainstream CNN architectures and clarify how these approaches overcome the difficulties of CNN training and promote the CNN development. The advantages and disadvantages are analyzed in details. We believe those details are important for the research of CNN based image contour detection.
Next, we discuss several state-of-the-art methods for image contour detection using CNNs with comprehensive analysis and discussions, with the goal to show the problems current state-of-the-art methods are trying to solve.
We discuss the challenges and significance of contour detection in medical images and review the historical approaches to solve these problems. Then we review the CNN based medical image contour detection and segmentation methods that leverage on recent advances in CNNs for contour detection and segmentation. Finally, we discuss the problems of existing methods and point out potential directions.

This paper attempts to cover necessary technical ingredients of state-of-the-art CNNs and connect their applications in the medical image domain. Compared with the various methods in the computer vision community, we point out the current diversity deficiency in the medical image contour detection and segmentation and provide potential directions and technical suggestions.

\section*{References}

\bibliographystyle{elsarticle-harv} 
\bibliography{egbib}

%
%
%
\end{document}